\documentclass[lettersize,journal]{IEEEtran}
\usepackage{amsmath,amsfonts}
\usepackage{algorithmic}
\usepackage{algorithm}
\usepackage{array}
\usepackage[caption=false,font=normalsize,labelfont=sf,textfont=sf]{subfig}
\usepackage{textcomp}
\usepackage{stfloats}
\usepackage{url}
\usepackage{verbatim}
\usepackage{graphicx}
\usepackage{cite}
\usepackage{kotex}
\usepackage{booktabs}
\usepackage{multirow}
\usepackage{tikz-cd}
\begin{document}
\bstctlcite{RALBSTcontrol}
\title{APG-Planner: Learning Vision-Based High-Speed Trajectory Planning via Analytic Policy Gradient}

\author{Jinwoo Lee$^{*}$, Geunsik Bae$^{*}$, Changseung Kim, Jiwoo Kim, Woojae Shin, and Hyondong Oh$^{\dagger}$ \thanks{$^{*}$Equal contribution. $^{\dagger}$Corresponding author.}}

\maketitle

\begin{abstract}
Vision-based high-speed navigation requires a trajectory planner that can reactively generate smooth, collision-free trajectories from onboard visual observations. Deep reinforcement learning offers a promising approach to vision-based trajectory planning, as it enables the optimization of sequential decision-making policies in complex environments. However, model-free policy gradient methods rely on sample-based policy gradient estimation, which yields noisy gradient estimates and inefficient training. To this end, we propose a framework for learning trajectory planning policies using the analytic policy gradient (APG) method and use it to train APG-Planner, a vision-based high-speed trajectory planner. The proposed framework leverages differential flatness to analytically reconstruct vehicle states from planned trajectories, thereby enabling differentiable trajectory-level state transitions without requiring an explicit low-level dynamics model. By iteratively performing trajectory planning and differentiable state transitions, we construct a differentiable episode rollout and analytically compute policy gradients from episode-level losses, improving both training efficiency and policy performance. After approximately 30 minutes of training on a single GPU, APG-Planner demonstrates comparable or superior performance to existing baseline methods in simulation and achieves speeds of up to 11\,m/s in real-world flight experiments. We further validate the proposed framework on a skid-steer ground robot, demonstrating its applicability to differentially flat robotic systems beyond quadrotors.
\end{abstract}


\section{Introduction}

Vision-based high-speed flight has been widely studied to fully exploit the agility of quadrotors. The central challenge is to plan collision-free trajectories online from visual observations under limited onboard computation. Conventional systems reconstruct a geometric representation of the environment and then perform a trajectory search or optimization~\cite{ego,wang2022gcopter,9592663}. Although effective, these approaches introduce computational latency that may limit rapid replanning for obstacle avoidance. To address this limitation, recent studies have proposed learning-based planners that directly generate trajectories from depth observations~\cite{loquercio2021highspeed,rapid,lpnet,10528860}.

Recent works on learning-based trajectory planning have adopted training paradigms such as imitation learning~\mbox{\cite{loquercio2021highspeed,rapid}} and guidance learning~\cite{10528860}. Imitation learning can offer advantages in sample efficiency but is limited by the availability and quality of expert trajectories. Guidance-based learning removes the need for expert demonstrations by exploiting gradients from a predefined planning-cost landscape, but such guidance remains local to each planning decision. 

In high-speed flight, each planned trajectory affects future states and decisions, influencing overall flight performance. Reinforcement learning (RL) offers a promising approach to this problem, as it optimizes sequential decisions with respect to task-level objectives. However, training model-free RL policies for high-speed flight remains challenging, as model-free policy gradient methods rely on sampled rollouts to estimate policy gradients. These estimates can exhibit high variance, making training sample-intensive and slowing convergence~\cite{newton}. An alternative should enable task-level optimization while providing more accurate policy gradients.

Analytic policy gradient (APG) methods provide such an alternative by constructing a differentiable episode rollout and backpropagating task-loss through it~\cite{apg_controller,brax}. Previous studies have employed APG to learn control policies, typically by propagating system states through differentiable dynamics. Recent work has successfully applied this approach to learn an end-to-end control policy for vision-based high-speed drone flight~\cite{newton}. However, this approach required a differentiable dynamics simulator calibrated using real-world flight data, necessitating recalibration of parameters such as aerodynamic drag across vehicles, while its end-to-end formulation limits interpretability. Therefore, learning high-speed trajectory planners remains a distinct and important research direction, providing explicit trajectories that facilitate interpretation and fallback design while integrating with existing low-level controllers to reduce reliance on low-level dynamics modeling.

In this work, we propose a learning framework that uses APG to train APG-Planner, a vision-based high-speed trajectory planner. The proposed policy outputs trajectory segments in the flat-output space, while an existing low-level controller tracks the planned trajectories. Training a trajectory planning policy with APG requires differentiable state transitions between successive planning steps. We construct these transitions directly from planned trajectories using differential flatness, instead of explicitly simulating low-level vehicle dynamics. 

Differential flatness enables quadrotor states to be represented in terms of position, heading, and a finite number of their derivatives~\cite{mellinger2011minsnap}, and has been widely exploited for trajectory optimization~\cite{wang2022gcopter,9592663}. We take this idea a step further by using differential flatness not only to parameterize trajectories, but also to define the state transition between successive planning decisions. Following a receding-horizon scheme, we analytically evaluate the planned trajectory at each replanning instant and reconstruct the corresponding vehicle state through the flatness map for the subsequent planning decision. This enables differentiable trajectory-level state transitions without explicitly modeling or unrolling low-level vehicle dynamics.

By iteratively performing trajectory planning and differentiable state transitions, we construct differentiable episode rollouts and directly compute policy gradients from episode-level losses. This allows the planner to account for the downstream effects of sequential planning decisions during policy optimization while avoiding sampling-based policy-gradient estimation. 
Furthermore, by leveraging the flatness map of the target system, our framework enables the training of trajectory planners without requiring platform-specific low-level dynamics models. Moreover, the proposed framework is not limited to quadrotors and can be extended to other differentially flat robotic systems.

The main contributions of this work are as follows:
\begin{itemize}
    \item We formulate differentiable trajectory-level state transitions using differential flatness, without explicitly modeling low-level system dynamics.

    \item Based on this formulation, we propose an APG-based framework for training trajectory planning policies. Using this framework, we train APG-Planner in just 30 minutes on a single GPU.

    \item We demonstrate the performance of APG-Planner through extensive experiments, achieving competitive results in simulation and peak flight speeds of 11\,m/s in real-world quadrotor flights. Further experiments on a skid-steer ground robot validate its applicability beyond quadrotors.
\end{itemize}

\section{Preliminaries}
\label{sec:preliminaries}
The proposed framework builds on two key concepts: differential flatness, which provides an analytic mapping from planned trajectories to vehicle states, and APG, which enables policy optimization through differentiable rollouts. Throughout this paper, \(t\) denotes continuous trajectory time, while \(k\) indexes discrete planning decisions.

\subsection{Differential Flatness}
\label{sec:diff_flat}
A system is differentially flat if its state and input can be written as algebraic functions of a set of flat outputs and a finite number of their time derivatives. For a nominal quadrotor model, the translational dynamics and attitude kinematics are
$$
\dot{\mathbf{p}}=\mathbf{v}, \qquad
m\dot{\mathbf{v}}=m\mathbf{g}+f\mathbf{R}\mathbf{e}_3, \qquad
\dot{\mathbf{R}}=\mathbf{R}[\boldsymbol{\omega}]_{\times},
$$
where \(\mathbf{p},\mathbf{v} \in \mathbb{R}^3\) , \(\mathbf{R}\in SO(3)\), and \(\boldsymbol{\omega}\in\mathbb{R}^3\) denote position, velocity, body-to-world rotation, and body angular velocity, respectively. Here, \(m\) is the mass, \(f\) is the collective thrust, \(\mathbf{g}\) is world-frame gravity, and \(\mathbf{e}_3=[0,0,1]^\top\).

The quadrotor is differentially flat with flat outputs
$$
\boldsymbol{\sigma}(t)=
\begin{bmatrix}
\mathbf{p}(t)^\top & \psi(t)
\end{bmatrix}^{\top},
$$
where \(\psi\) is a heading angle. Position, velocity, and acceleration are obtained directly from \(\mathbf{p}(t)\), \(\dot{\mathbf{p}}(t)\), and \(\ddot{\mathbf{p}}(t)\), respectively. Since thrust acts along the body \(z\)-axis, the required acceleration determines the direction of that axis:
$$
\mathbf{b}_3(t)=
\frac{\ddot{\mathbf{p}}(t)-\mathbf{g}}
{\|\ddot{\mathbf{p}}(t)-\mathbf{g}\|_2}.
$$
The heading angle defines the horizontal reference vector
$$
\mathbf{b}_{1c}(t)=
\begin{bmatrix}
\cos\psi(t) & \sin\psi(t) & 0
\end{bmatrix}^{\top}.
$$
The remaining body axes and attitude matrix are recovered as
$$
\mathbf{b}_2(t)=
\frac{\mathbf{b}_3(t)\times\mathbf{b}_{1c}(t)}
{\|\mathbf{b}_3(t)\times\mathbf{b}_{1c}(t)\|_2},
\quad
\mathbf{b}_1(t)=\mathbf{b}_2(t)\times\mathbf{b}_3(t),
$$
$$
\mathbf{R}(t)=
\begin{bmatrix}
\mathbf{b}_1(t) & \mathbf{b}_2(t) & \mathbf{b}_3(t)
\end{bmatrix}.
$$
The reconstruction of higher-order quantities, such as the body angular velocity \(\boldsymbol{\omega}(t)\), is described in \cite{mellinger2011minsnap}. Collectively, these relations define the analytic flatness map from the flat outputs and their derivatives to the state $\mathbf{x}_t$: 

$$
\mathbf{x}_t=\Phi_\mathbf{x}\left(
\boldsymbol{\sigma}(t),
\dot{\boldsymbol{\sigma}}(t),
\ddot{\boldsymbol{\sigma}}(t),\ldots
\right).
$$

The flatness map thus recovers nominal states without numerically
integrating the dynamics. In the proposed framework, we leverage this property to construct differentiable state transitions without explicitly modeling low-level vehicle dynamics, enabling APG-based policy optimization.

\subsection{Analytic Policy Gradients}
\label{sec:prelim_apg}

Consider a policy $\pi_\theta$ parameterized by $\theta$ and a
state-transition model $F$. At decision step $k$, the policy generates
an action from the current observation $\mathbf{o}_k$, and the model
determines the next state:
\begin{equation*}
    \mathbf{a}_k=\pi_\theta(\mathbf{o}_k), \qquad
    \mathbf{x}_{k+1}=F(\mathbf{x}_k,\mathbf{a}_k).
\end{equation*}
Let $J_k(\mathbf{x}_k,\mathbf{a}_k)$ denote the task loss at step $k$.
Over an episode of $K$ steps, the objective is
\begin{equation*}
    \mathcal{L}
    =\frac{1}{K}\sum_{k=0}^{K-1}
    J_k(\mathbf{x}_k,\mathbf{a}_k).
\end{equation*}

Optimizing this objective requires computing the policy gradient $\nabla_{\theta}\mathcal{L}$. Model-free policy-gradient methods, such as proximal policy optimization (PPO), estimate policy gradients from sampled experience without differentiating through state transitions~\cite{schulman2017ppo}. In contrast, APG leverages differentiable state transitions and loss functions to construct an entire episode as a single differentiable computation graph through rollouts, enabling direct computation of policy gradients via backpropagation~\cite{newton}. By accounting for the effects of each policy action on both immediate and future losses, backpropagation through time (BPTT) yields the following gradient, assuming a parameter-independent initial state:

\begin{equation*}
 \begin{aligned}
 \nabla_{\theta}\mathcal{L}
 ={}&\frac{1}{K}\sum_{k=0}^{K-1}
 \Bigg[
 \underbrace{
  \frac{\partial J_k}{\partial\mathbf{a}_k}
  \frac{\partial\mathbf{a}_k}{\partial\theta}
 }_{\text{local gradient}}
 +\underbrace{
  \sum_{i=1}^{k}
  \frac{dJ_k}{d\mathbf{x}_k}
  \left(
   \prod_{j=i+1}^{k}\!
   \frac{\partial\mathbf{x}_j}{\partial\mathbf{x}_{j-1}}\!\!
  \right)\!
  \frac{\partial\mathbf{x}_i}{\partial\theta}
 }_{\text{temporal gradient}}
 \Bigg].
 \end{aligned}
\end{equation*}
The local term represents the direct effect of each policy action on the current loss, while the temporal term captures its effect on future losses through state transitions. Since actions depend on the current state, state changes affect both the next state and current loss through the policy. These dependencies are expressed using the chain rule:
\begin{equation*}
 \begin{aligned}
 \frac{\partial\mathbf{x}_j}{\partial\mathbf{x}_{j-1}}
 &=\frac{\partial F}{\partial\mathbf{x}}
 +\frac{\partial F}{\partial\mathbf{a}}
 \frac{\partial\mathbf{a}_{j-1}}{\partial\mathbf{x}_{j-1}},\\
 \frac{dJ_k}{d\mathbf{x}_k}
 &=\frac{\partial J_k}{\partial\mathbf{x}_k}
 +\frac{\partial J_k}{\partial\mathbf{a}_k}
 \frac{\partial\mathbf{a}_k}{\partial\mathbf{x}_k}.
 \end{aligned}
\end{equation*}

The policy parameters are updated as $
\theta\leftarrow\theta-\eta\nabla_\theta\mathcal{L}$, where $\eta$ is a learning rate. In the proposed framework, we leverage APG to learn a high-speed trajectory planning policy.

\section{Methodology}
\label{sec:method}

This section presents the proposed framework for learning a vision-based high-speed trajectory planner. We first describe the policy architecture and trajectory generation method. We then explain how differential flatness-based state transitions are used to construct differentiable episode rollouts through receding-horizon planning. Finally, we introduce the loss function and APG-based policy optimization, followed by the training setup and deployment procedure.

\begin{figure}[tbp]
 \centering
 \includegraphics[width=0.9\linewidth,trim={300 20 300 0},clip]
 {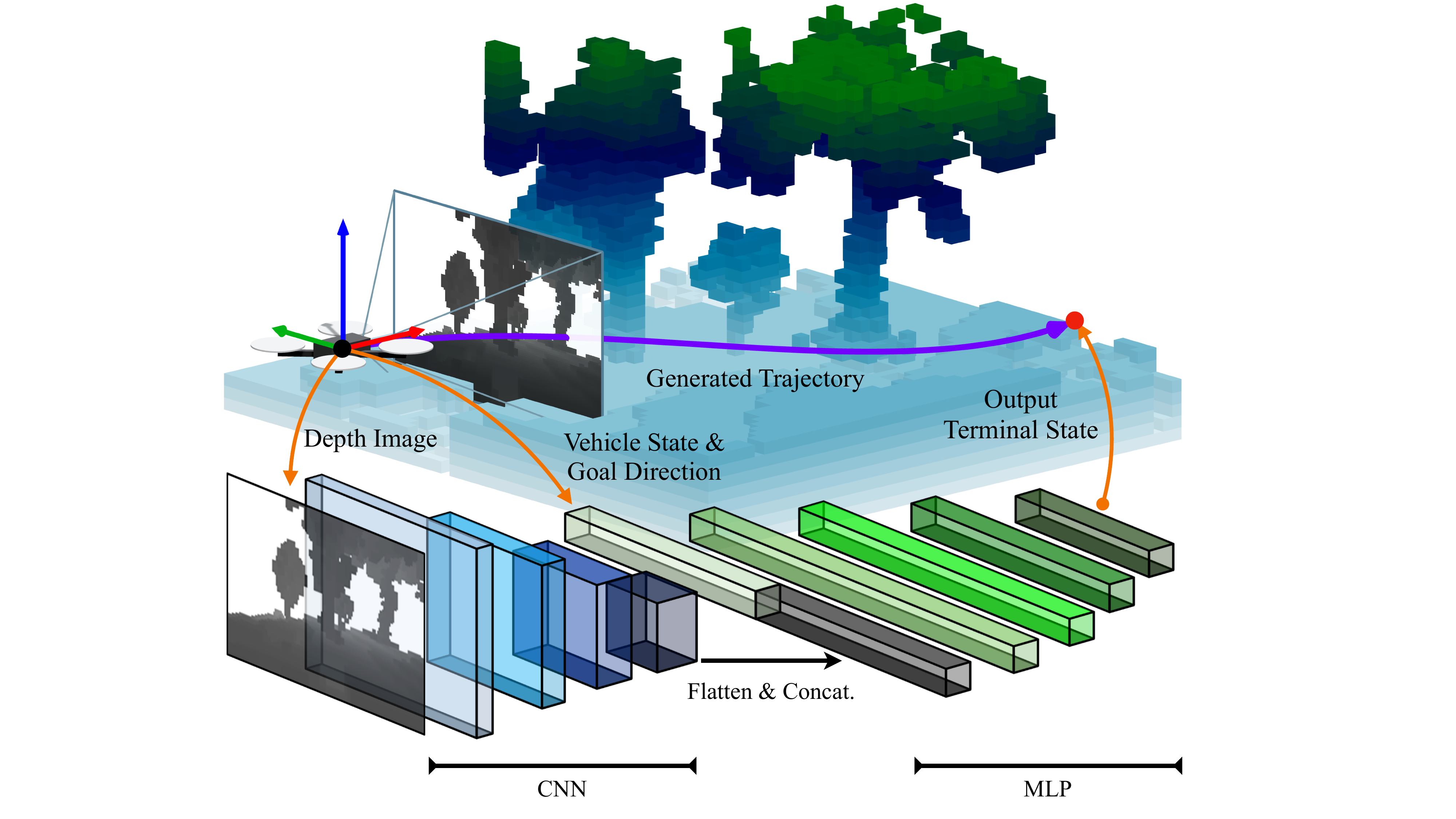}
 \caption{Policy inference (bottom) and the resulting trajectory (top). The network outputs terminal conditions, from which the trajectory is constructed.}
 \label{fig:network_inference}
 \vspace{-4mm}
\end{figure}
\subsection{Trajectory Planner Architecture}
\label{sec:traj_generation}

\subsubsection{Policy Network}

At planning step $k$, the policy receives a depth image
$I_k\in\mathbb{R}^{H\times W}$, vehicle-state features $\mathbf{s}_k$,
and a goal vector $\mathbf{d}_k^{\mathcal{V}}$, where $H$ and $W$ denote the image
height and width, respectively.
The superscript $\mathcal{V}$ denotes a yaw-aligned frame that shares the
world vertical axis and rotates with the vehicle's current yaw.
All other vectors are expressed in the world frame $\mathcal{W}$.
The state features are $\mathbf{s}_k=[(\mathbf{v}_k^{\mathcal{V}})^\top, (\boldsymbol{\alpha}_k^{\mathcal{V}})^\top,\phi_k,\vartheta_k]^\top \in\mathbb{R}^{8}$, where $\mathbf{v}$ and $\boldsymbol{\alpha}$ denote
translational velocity and acceleration, while $\phi$ and $\vartheta$ are roll and pitch, respectively. The goal vector $\mathbf{d}_k^{\mathcal{V}}\in\mathbb{R}^{3}$
points toward the goal, with its magnitude adjusted according to
the remaining distance. Its magnitude stays at one when the goal
is farther than $\ell_g$ and decreases linearly as the vehicle
approaches the goal within this distance, allowing the policy
to capture both the goal direction and proximity.

As illustrated in Fig.~\ref{fig:network_inference}, a convolutional neural
network (CNN) encodes $I_k$. Its flattened feature is concatenated with
$\mathbf{s}_k$ and $\mathbf{d}_k^{\mathcal{V}}$ and processed by a multilayer
perceptron (MLP) to output an action $\mathbf{a}_k$, in a form of terminal conditions for the trajectory to be generated:
\begin{equation*}
 \begin{aligned}
 \mathbf{a}_k&=\pi_\theta(I_k,\mathbf{s}_k,\mathbf{d}_k^{\mathcal{V}})
 =
 \begin{bmatrix}
  (\Delta\mathbf{p}_T^{\mathcal{V}})^\top
  (\mathbf{v}_T^{\mathcal{V}})^\top
  (\boldsymbol{\alpha}_T^{\mathcal{V}})^\top
 \end{bmatrix}^{\top}\in\mathbb{R}^{9}.
 \end{aligned}
\end{equation*}
Each output component is bounded by the corresponding entry of
$\Delta\mathbf{p}_{\max}$, $\mathbf{v}_{\max}$, or
$\boldsymbol{\alpha}_{\max}$. We use sigmoid outputs for forward
displacement and velocity, and
hyperbolic-tangent outputs for the remaining components.

\subsubsection{Trajectory Generation}

A trajectory segment is constructed from the current state and the terminal conditions output by the policy. Trajectory generation is performed in the world frame $\mathcal{W}$. Let $\mathbf{Q}_k$ denote the rotation from the current yaw-aligned frame $\mathcal{V}$ to $\mathcal{W}$. The terminal conditions are transformed as
\begin{equation*}
 \begin{aligned}
 \mathbf{p}_T&=\mathbf{p}_k
              +\mathbf{Q}_k\,\Delta\mathbf{p}_T^{\mathcal{V}}, \quad
 \mathbf{v}_T=\mathbf{Q}_k\,\mathbf{v}_T^{\mathcal{V}},\quad
 \boldsymbol{\alpha}_T=\mathbf{Q}_k\,\boldsymbol{\alpha}_T^{\mathcal{V}}.
 \end{aligned}
\end{equation*}
We represent the position trajectory by a quintic polynomial:
\begin{equation*}
\mathbf{p}_k(t)\!
=\!
\mathbf{c}_{k_0}\!
+\mathbf{c}_{k_1}t\!
+\mathbf{c}_{k_2}t^2\!
+\mathbf{c}_{k_3}t^3\!
+\mathbf{c}_{k_4}t^4\!
+\mathbf{c}_{k_5}t^5,
\,\, t\in[0,T].
\end{equation*}
For fixed trajectory time duration $T$, the current state and transformed terminal conditions are used as boundary conditions to uniquely determine its coefficients:
\begin{equation*}
\begin{aligned}
\mathbf{p}_k(0)&=\mathbf{p}_k, \quad
\dot{\mathbf{p}}_k(0)=\mathbf{v}_k, \quad
\ddot{\mathbf{p}}_k(0)=\boldsymbol{\alpha}_k,\\
\mathbf{p}_k(T)&=\mathbf{p}_T, \quad
\dot{\mathbf{p}}_k(T)=\mathbf{v}_T, \quad
\ddot{\mathbf{p}}_k(T)=\boldsymbol{\alpha}_T.
\end{aligned}
\end{equation*}
Equivalently, the coefficients satisfy
\begingroup
\setlength{\arraycolsep}{3pt}
\begin{equation}
 \begin{bmatrix}
 1&0&0&0&0&0\\
 0&1&0&0&0&0\\
 0&0&2&0&0&0\\
 1&T&T^2&T^3&T^4&T^5\\
 0&1&2T&3T^2&4T^3&5T^4\\
 0&0&2&6T&12T^2&20T^3
 \end{bmatrix}
 \begin{bmatrix}
 {\mathbf{c}_{k_0}}\!\!^\top\\
 {\mathbf{c}_{k_1}}\!\!^\top\\
 {\mathbf{c}_{k_2}}\!\!^\top\\
 {\mathbf{c}_{k_3}}\!\!^\top\\
 {\mathbf{c}_{k_4}}\!\!^\top\\
 {\mathbf{c}_{k_5}}\!\!^\top
 \end{bmatrix}
 =
 \begin{bmatrix}
 {\mathbf{p}_k}\!\!^\top\\
 {\mathbf{v}_k}\!\!^\top\\
 {\boldsymbol{\alpha}_k}\!\!^\top\\
 {\mathbf{p}_T}\!\!^\top\\
 {\mathbf{v}_T}\!\!^\top\\
 {\boldsymbol{\alpha}_T}\!\!^\top
 \end{bmatrix}.
 \label{eq:boundary_matrix}
\end{equation}
\endgroup
Since $T$ is fixed, the polynomial coefficients are obtained by solving
the linear system in Eq.~\eqref{eq:boundary_matrix}. This yields the
position trajectory as a differentiable linear function of the boundary
conditions. The heading angle is then defined from the horizontal tangent
of the trajectory:
\begin{equation*}
\psi_k(t)
=\operatorname{atan2}\bigl(
\dot p_{k,y}(t),\dot p_{k,x}(t)\bigr).
\end{equation*}
When the horizontal speed falls below 0.1 m/s, where the tangent direction is ill-defined, the heading from the previous planning step is retained. The resulting position trajectory and heading angle together define the flat-output trajectory:
\begin{equation}
\boldsymbol{\sigma}_k(t)=
\begin{bmatrix}
\mathbf{p}_k(t)^\top & \psi_k(t)
\end{bmatrix}^{\top},
\qquad t\in[0,T].
\label{eq:flat_output_trajectory}
\end{equation}
The current state $\mathbf{x}_k$ and policy action $\mathbf{a}_k$ therefore determine
the complete trajectory segment analytically, which is used for loss
evaluation and state transition.

\subsection{Receding-Horizon Episode Rollout}
\label{sec:learning_framework}

\begin{figure}[t]
 \centering
 \includegraphics[width=0.9\linewidth, trim= 450 200 450 200]{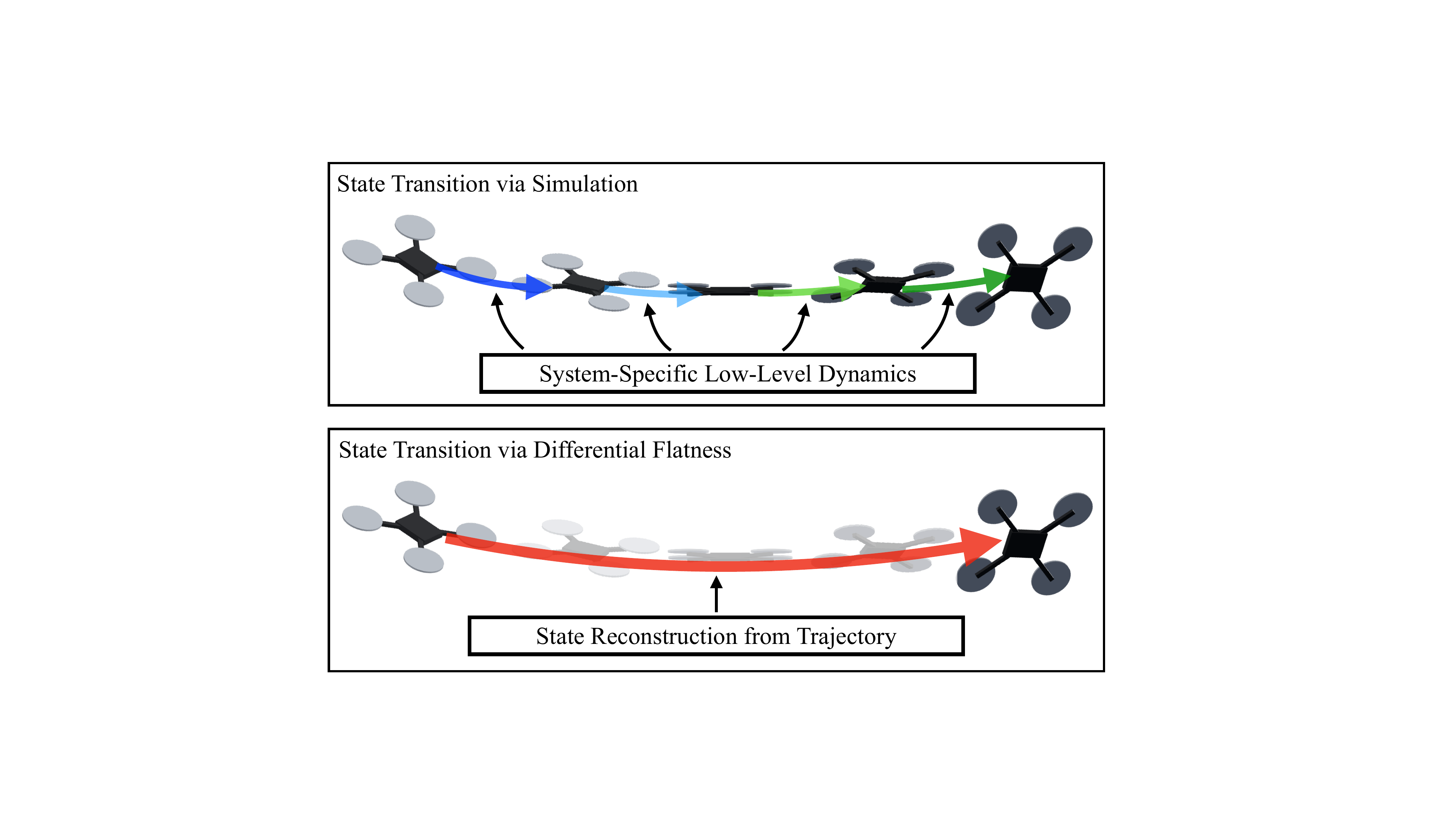}
 \caption{Comparison of state-transition models. Top: simulation advances
 the state through hardware-specific physics and actuator dynamics.
 Bottom: the transition in Eq.~\eqref{eq:transition} evaluates the
 trajectory at $\Delta T$ and reconstructs the nominal planning state
 through the flatness map.}
 \label{fig:flatness_transition}
\end{figure}

To construct a receding-horizon episode rollout from the trajectory segments generated above, the vehicle state must be propagated along each planned trajectory to provide the initial state for the next planning step. To this end, we formulate a state-transition model based on differential flatness.

Let $\mathbf{x}_k$ denote the vehicle state at planning step $k$. Using the differential-flatness relations described in Section~\ref{sec:diff_flat}, the nominal vehicle state at any time $t\in[0, T]$ along the planned trajectory can be analytically reconstructed from the flat-output trajectory $\boldsymbol{\sigma}_k(t)$ as
\begin{equation*}
\hat{\mathbf{x}}_k(t)
=\Phi_\mathbf{x}\bigl(
\boldsymbol{\sigma}_k(t),
\dot{\boldsymbol{\sigma}}_k(t),
\ddot{\boldsymbol{\sigma}}_k(t),\ldots
\bigr).
\end{equation*}

In the receding-horizon rollout, the state reconstructed at $t=\Delta T$ is used as the initial state for the next planning step. The resulting state transition is defined as
\begin{equation}
\mathbf{x}_{k+1}
=F(\mathbf{x}_k,\mathbf{a}_k)
:=\hat{\mathbf{x}}_k(\Delta T).
\label{eq:transition}
\end{equation}

As illustrated in Fig.~\ref{fig:flatness_transition}, this formulation directly connects successive planning decisions at the trajectory level without integrating the controller or introducing an explicit low-level vehicle dynamics model.

At each planning step, the policy generates an action $\mathbf{a}_k$ from the current observation. The action, together with the current state $\mathbf{x}_k$, determines a trajectory segment, from which the next planning state $\mathbf{x}_{k+1}$ is reconstructed at $t=\Delta T$. The next depth image $I_{k+1}$ is then rendered using the reconstructed position and attitude, and combined with the updated state feature vector $\mathbf{s}_{k+1}$ and goal vector $\mathbf{d}_{k+1}^{\mathcal V}$ to form input to the next policy decision. Repeating this process over $K$ planning steps constructs a differentiable episode rollout.

\subsection{Loss Function}
\label{sec:loss}

To evaluate the policy over the state-action sequence generated by
the receding-horizon rollout and provide a learning signal for policy
optimization, we define a loss function tailored to high-speed
autonomous flight. Specifically, the loss considers three objectives:
goal progress, obstacle avoidance, and trajectory smoothness.
The episode objective is defined as the average of the per-step
trajectory loss $J_k$ over $K$ replanning decisions:
\begin{equation}
 \mathcal{L}=\frac{1}{K}\sum_{k=0}^{K-1}
 J_k(\mathbf{x}_k,\mathbf{a}_k).
 \label{eq:total_loss}
\end{equation}
Goal progress is evaluated at the visited position $\mathbf{p}_k$,
while obstacle avoidance and smoothness are evaluated over the full
planned horizon $T$. Although only the first $\Delta T$ is executed,
the remainder of the segment is still evaluated as the policy's intended
continuation. 
The goal and jerk costs at step $k$ are
\begin{equation*}
\begin{aligned}
J_{\mathrm{goal},k}
&=\omega_k
\frac{\|\mathbf{p}_k-\mathbf{p}_{\mathrm{goal}}\|_2}{g_0}, \quad 
J_{\mathrm{jerk},k}
=\frac{1}{T}\int_0^T
\|\dddot{\mathbf{p}}_k(t)\|_2^2\,dt.
\end{aligned}
\end{equation*}
Here, $g_0$ is the initial goal distance, and $\omega_k$ linearly
increases with episode progress up to a factor of 10, placing greater emphasis on goal distance at later planning steps. The jerk cost is evaluated in closed form for the quintic trajectory.

For obstacle avoidance, we sample $N$ positions along the planned
trajectory and compute the clearance $d_i$ to the nearest obstacle.
The obstacle cost is defined as
\begin{equation*}
J_{\mathrm{obs},k}
=\frac{1}{N}\sum_{i=1}^{N}w_i\,\ell(d_i),
\end{equation*}
where
\begin{equation*}
\ell(d)
=\max(0,\rho-d)^2
+\lambda_{\mathrm{pen}}\operatorname{softplus}(-\kappa d)
\end{equation*}
penalizes proximity to the safety boundary and boundary violations. We use \(\rho=1\) m, \(\kappa=32\) m\(^{-1}\), and \(\lambda_{\mathrm{pen}}=2.5\), which set the proximity margin, sharpness, and weight of the penetration penalty, respectively. The sample weight is defined as
\begin{equation*}
w_i
=\max\left\{
1,\,
\frac{d_i^{\mathrm{prev}}-d_i}
{v_{\mathrm{ref}}\Delta T}
\right\},
\end{equation*}
which emphasizes samples whose clearance decreases relative to the
corresponding sample in the previous plan. Here, $v_{\mathrm{ref}}$ is
a reference velocity used to normalize the clearance decrease.

Overall, the loss for policy updates is formulated as a weighted sum of the three cost terms:
\begin{equation*}
J_k
=
\lambda_{\mathrm{goal}} J_{\mathrm{goal},k}
+
\lambda_{\mathrm{jerk}} J_{\mathrm{jerk},k}
+
\lambda_{\mathrm{obs}} J_{\mathrm{obs},k},
\end{equation*}
where $\lambda_{\mathrm{goal}}$, $\lambda_{\mathrm{jerk}}$, and $\lambda_{\mathrm{obs}}$ are the weights for the goal, jerk, and obstacle avoidance costs, respectively.

\subsection{Policy Optimization}
\label{sec:apg}

\begin{figure*}[t]
 \centering
 \includegraphics[width=0.99\textwidth,trim={120 230 120 230},clip]
 {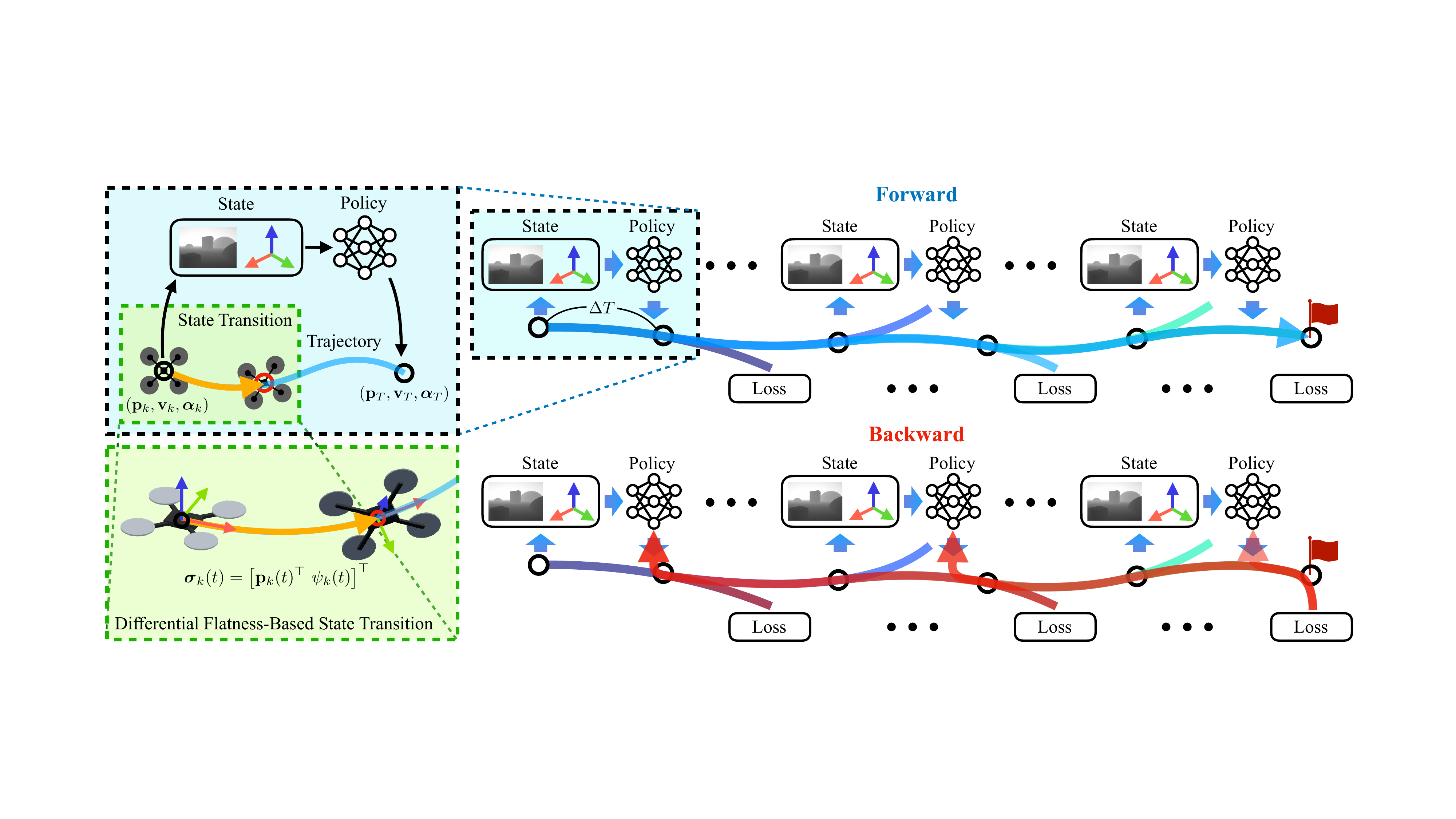}
 \caption{Overview of the proposed learning framework. At each step, the policy generates a trajectory segment of duration $T$, and the rollout advances by $\Delta T$ using the proposed state transition model based on differential flatness. The loss is computed at each step, and this process is repeated in a receding-horizon manner to form an episode. The policy is optimized using BPTT over the episode.}
 \label{fig:network}
 \vspace{-4mm}
\end{figure*}
The episode objective defined in Eq.~\eqref{eq:total_loss}
evaluates the sequence of planning decisions generated by the
receding-horizon rollout. Since the rollout is differentiable
through the proposed state-transition model, gradients of this
objective can be propagated across successive planning steps.
We therefore optimize the policy using APG with BPTT through the state transitions defined in
Eq.~\eqref{eq:transition}. Figure~\ref{fig:network}
summarizes the training framework. To improve gradient stability,
temporal gradients between successive planning decisions are propagated
only through position, velocity, and acceleration. The reconstructed
attitude and rendered depth image are used to update the forward
observations but detached from backpropagation, thereby simplifying the
temporal gradient path through the rollout.

Following~\cite{newton}, we attenuate long-range gradients by multiplying
the state-transition Jacobians in the temporal product by a decay factor
$\gamma$. The resulting BPTT update is given by
\begin{equation*}
 \begin{aligned}
 \widetilde{\nabla}_{\theta}\mathcal{L}
 ={}&\frac{1}{K}\sum_{k=0}^{K-1}\!
 \Bigg[
 \underbrace{
  \frac{\partial J_k}{\partial\mathbf{a}_k}
  \frac{\partial\mathbf{a}_k}{\partial\theta}
 }_{\text{local gradient}}
 \!+\!\underbrace{
  \sum_{i=1}^{k}
  \frac{dJ_k}{d\mathbf{x}_k}
  \left(
   \prod_{j=i+1}^{k}\!\!
   \gamma\frac{\partial\mathbf{x}_j}
                  {\partial\mathbf{x}_{j-1}}\!\!
  \right)\!\!
  \frac{\partial\mathbf{x}_i}{\partial\theta}
 }_{\text{temporal gradient}}
 \Bigg].
 \end{aligned}
\end{equation*}
Using the derivative conventions in Section~\ref{sec:prelim_apg}, each
transition Jacobian includes the policy's retained state-dependent inputs,
trajectory generation, and evaluation at $\Delta T$. The derivative
$dJ_k/d\mathbf{x}_k$ combines the direct path through the initial boundary
conditions of $\boldsymbol{\sigma}_k$ and the path through
$\mathbf{a}_k$.

\subsection{Training Setup and Deployment}

The policy was trained using the proposed framework with the parameters summarized in Table~\ref{tab:training_setup}. To improve generalization across diverse obstacle environments and flight distances, we applied several forms of domain randomization. At the start of each episode, obstacle configurations, environment scales, start and goal positions, and global yaw were randomized. The environments included diverse obstacle structures, such as cylinders, Perlin-noise structures, and trees. Depth images were clipped and normalized over the specified range and then converted into inverse-depth values. Gaussian noise was applied to the depth images to improve robustness to observation variations. Training took approximately 30 minutes on an NVIDIA RTX 4080 SUPER GPU.

\begin{table}[!t]
 \centering
 \caption{Policy and training settings}
 \label{tab:training_setup}
 \footnotesize
 \setlength{\tabcolsep}{4pt}
 \renewcommand{\arraystretch}{1.2}
 \begin{tabular}{@{}>{\raggedright\arraybackslash}p{0.36\linewidth}
                    >{\raggedright\arraybackslash}p{0.60\linewidth}@{}}
 \toprule
 Parameter & Setting\\
 \midrule
Terminal bounds $(x,y,z)$ &
$\Delta\mathbf{p}_T^{\mathcal V}$:
$([0,20],[-10,10],[-10,10])$~m;\\
&
$\mathbf{v}_T^{\mathcal V}$:
$([0,20],[-10,10],[-10,10])$~m/s;\\
&
$\boldsymbol{\alpha}_T^{\mathcal V}$:
$[-10,10]$~m/s$^2$ (all axes)\\

Depth input &
$40\times64$ pixels, clipped to $[0.3,20]$~m\\

Goal-vector threshold &
$\ell_g=10$~m\\

% Trajectory / replanning &
% $T=1$~s, $\Delta T=0.1$~s\\
Trajectory duration & $T=1$~s\\
Replanning interval & $\Delta T=0.1$~s\\
Obstacle-cost samples &
$N=10$\\
Episode horizon &
$K=50$\\

Optimizer / learning rate &
AdamW / $10^{-3}$\\
Training configuration &
$10{,}000$ iterations, 64 parallel environments\\
Gradient decay &
$\gamma=0.98$\\
Loss weights &
$\lambda_{\mathrm{goal}}=7.5$,
$\lambda_{\mathrm{obs}}=3.0$,
$\lambda_{\mathrm{jerk}}=0.005$\\
 \bottomrule
 \end{tabular}
\end{table}

During deployment, the planner operates in a receding-horizon manner, with the trained policy generating a local trajectory from the latest state estimate, depth observation, and goal. For the quadrotor, a geometric controller~\cite{lee2010geometric} tracks the generated trajectory in a separate control loop, while the planner replans at intervals of~$\Delta T$.

\section{Simulation Study}
\label{sec:simulation}

To assess the benefits of the proposed method for high-speed trajectory planning, we conduct simulation experiments. We compare the flight performance of the learned planner with that of baseline planners under identical and repeatable conditions. We also conduct an ablation study to examine how temporal gradient propagation and the policy optimization method affect learning efficiency and policy performance.

\subsection{Benchmark Evaluation}
\label{sec:benchmark}

We compare APG-Planner with representative online  trajectory planning algorithms: EGO~\cite{ego}, AGILE~\cite{loquercio2021highspeed}, and YOPO~\cite{10528860}, which employ optimization, imitation learning, and guidance learning, respectively. We evaluate the planners across a range of flight speeds using three metrics: success rate, trajectory snap, and tracking error. Success rate is the fraction of trials that reach the goal without collision. Trajectory snap reflects smoothness and control demand, while tracking error measures the discrepancy between planned and executed trajectories.

We use the Flightmare simulator~\cite{flightmare} with a geometric controller for trajectory tracking. The quadrotor has a mass of $0.73$~kg, an arm length of $0.17$~m, and a maximum thrust-to-weight ratio of $3.35$. Depth observations have a $90^{\circ}$ horizontal field of view and a maximum range of $20$~m, and are downsampled via max pooling to match the training resolution.

\begin{figure}[t]
    \centering
    \includegraphics[
        width=0.98\linewidth,
        trim={0 50 0 0},
        clip
    ]{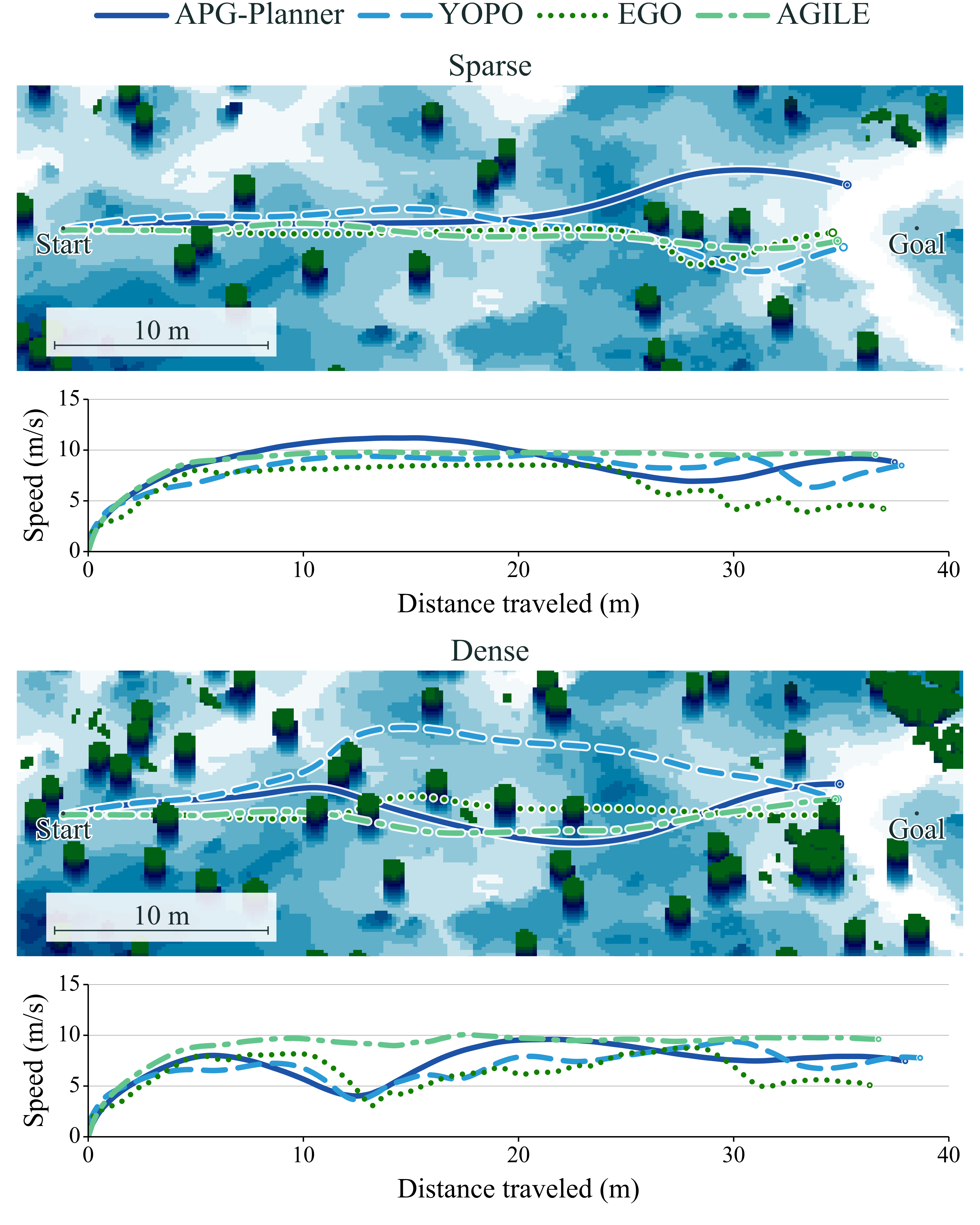}
    \caption{Flight trajectories generated by the baseline planners and the proposed planner in sparse and dense simulation environments. The proposed planner generates smoother obstacle-avoidance trajectories.}
    \label{fig:bench_vis}
    \vspace{-3mm}
\end{figure}

We evaluate all planners on the same 50 forest maps per tree density level: $1/16$~m$^{-2}$ (dense) and $1/25$~m$^{-2}$ (sparse), as shown in Fig.~\ref{fig:bench_vis}. Each trial involves a $40$~m course toward a fixed goal. A trial is considered successful if the vehicle reaches within $4$~m of the goal without collision. Initial conditions and collision criteria are identical across methods.

We use the code released by the respective authors for all baselines and the released network weights for learning-based methods. For YOPO and APG-Planner, we use YOPO's speed-adjustment mechanism, modifying the trajectory duration and terminal conditions while preserving the trajectory topology. For EGO, we adjust the maximum speed, while for AGILE, we use the speed-adjustment mechanism implemented in its publicly released codebase. Because these mechanisms differ across planners, the same speed setting does not yield the same actual flight speed. We therefore report performance against the mean flight speed measured in each trial rather than against the commanded target speed.

\begin{figure}[t]
    \centering
    \includegraphics[
        width=0.98\linewidth,
        trim={50 10 40 80},
        clip
    ]{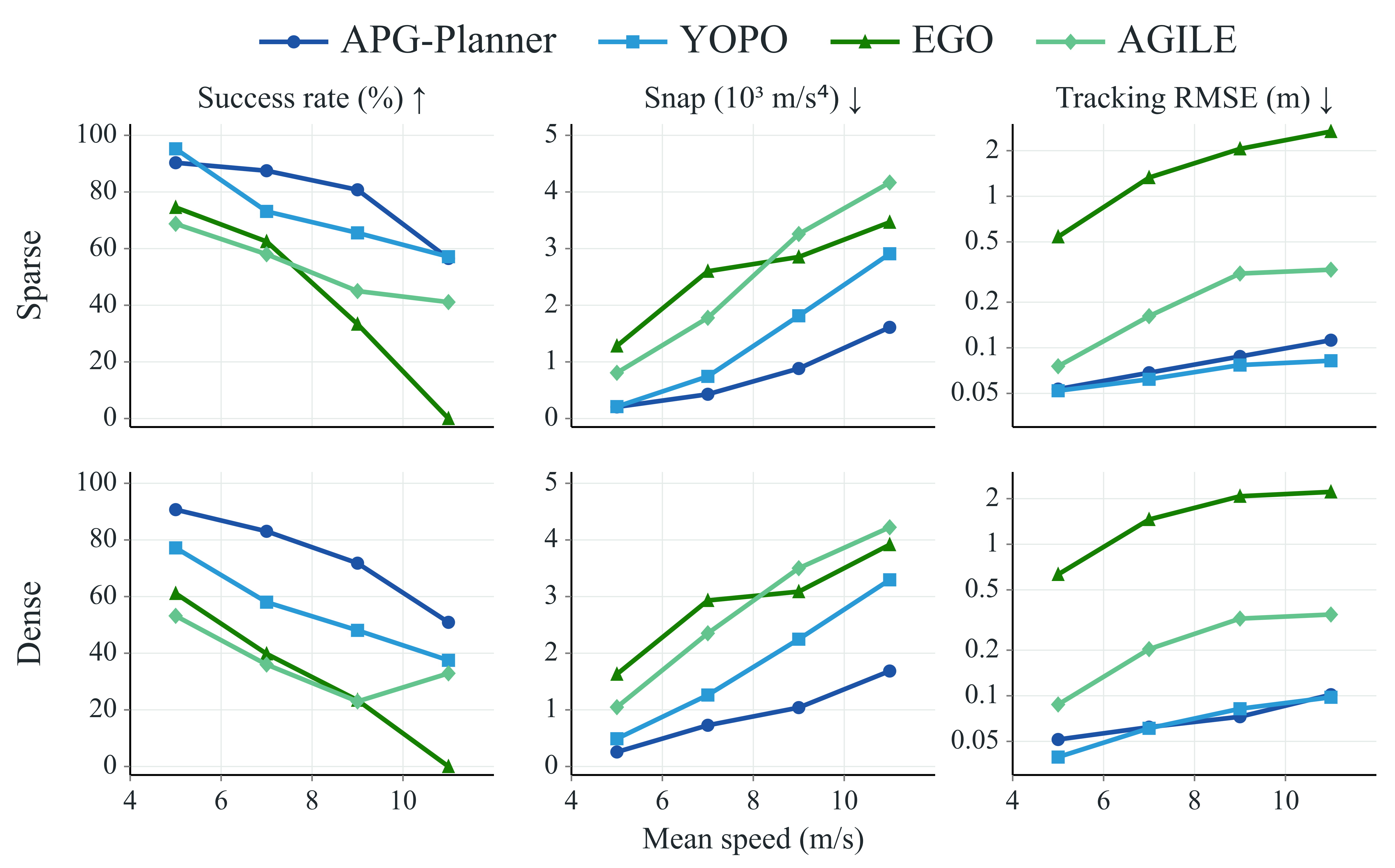}
    \caption{Success rate, trajectory snap, and tracking RMSE as functions of flight speed for the compared planners in sparse and dense obstacle environments.}
    \label{fig:bench_all}
\end{figure}

As shown in Fig.~\ref{fig:bench_all}, the proposed method achieves competitive or higher success rates in both sparse and dense environments, while generating trajectories with lower snap than the baselines, particularly at higher speeds. These results indicate improved obstacle avoidance performance together with smoother motion.

These performance differences can be interpreted in terms of the trajectory generation and learning strategies of each method. EGO relies on online gradient-based optimization and may struggle to obtain smooth, collision-free trajectories within the available replanning time during high-speed flight. In contrast, AGILE and YOPO use neural network inference to achieve rapid replanning and low tracking error. However, AGILE learns its trajectory-generation policy through imitation learning, making its performance limited by the quality and coverage of expert demonstrations. YOPO mitigates this issue through guidance learning based on trajectory costs, but does not explicitly account for long-term costs across successive replanning steps, which may limit its performance.

APG-Planner uses temporal gradient propagation during training to account for how current decisions affect subsequent replanning steps. This helps explain its ability to maintain high success rates and trajectory smoothness even at high speeds. It also achieves low tracking error in evaluations incorporating full quadrotor dynamics and a geometric controller, demonstrating that the generated avoidance trajectories can be accurately tracked during high-speed flight.

\subsection{Ablation Study}

We conduct two analyses to examine how APG-based training affects planning behavior and learning efficiency. We first qualitatively compare the planners' behavior in a wall-avoidance scenario, then examine the effects of temporal gradient propagation and the policy optimization method.

\begin{figure}[t]
    \centering
    \includegraphics[
        width=0.98\linewidth,
        trim={145 50 180 200},
        clip
    ]{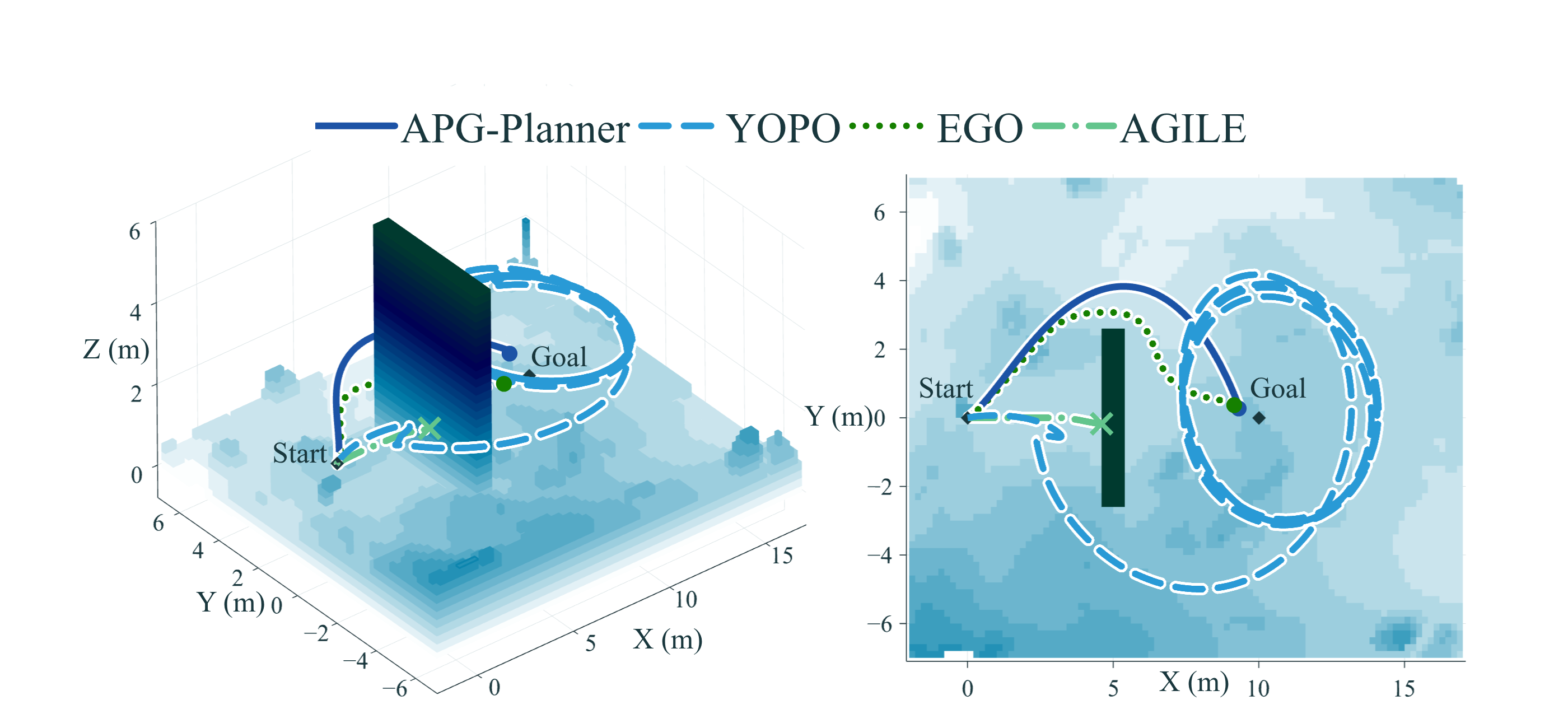}
    \caption{Comparison of flight trajectories in a wall-avoidance scenario. AGILE collides with the wall, and YOPO turns back. EGO initiates avoidance close to the wall, whereas APG-Planner begins turning earlier and follows a smooth path around the obstacle to reach the goal.}
    \label{fig:wall}
    \vspace{-3mm}
\end{figure}

\subsubsection{Wall-Avoidance Behavior}

We construct a scenario with a wall-shaped obstacle between the start and goal. Smoothly reaching the goal in this environment requires initiating a turn before approaching the wall closely and following a detour over multiple replanning steps.

As shown in Fig.~\ref{fig:wall}, AGILE collides with the wall, while YOPO turns back in front of it. Both methods use training signals for individual trajectory segments but do not directly optimize the long-term effects of current decisions. This limitation can lead to collisions or excessive detours around walls. 
YOPO also tends to circle around the goal, possibly trapped in a local minimum by favoring short-term smoothness over long-term goal progress.

EGO generates a collision-free trajectory through online gradient-based optimization and reaches the goal, but the reference trajectory lacks smoothness after the avoidance maneuver. In contrast, APG-Planner uses temporal gradient propagation during training to account for how current decisions affect subsequent flight behavior. By considering these long-term effects, it initiates its turn earlier and follows a smooth path around the wall to reach the goal.

\subsubsection{Policy Optimization Methods}

We compare policy optimization methods using the same training setup described in Section~\ref{sec:method}, varying only the optimization method to assess its effects on learning dynamics and the resulting flight behavior. We evaluate full APG, APG without temporal gradient, which blocks gradient propagation across replanning steps, and PPO, a representative model-free policy-gradient method.

APG without temporal gradient blocks gradients across replanning steps while retaining them within each step. The goal loss is evaluated at $t=\Delta T$ along each generated trajectory to provide a learning signal for goal progress. PPO is implemented using Stable-Baselines3~\cite{sb3}, with hyperparameters tuned for stable training. Its reward is the negative of the APG task loss to align their objectives.

\begin{figure}[t]
    \centering
    \includegraphics[
        width=0.98\linewidth,
        trim={0 0 0 0},
        clip
    ]{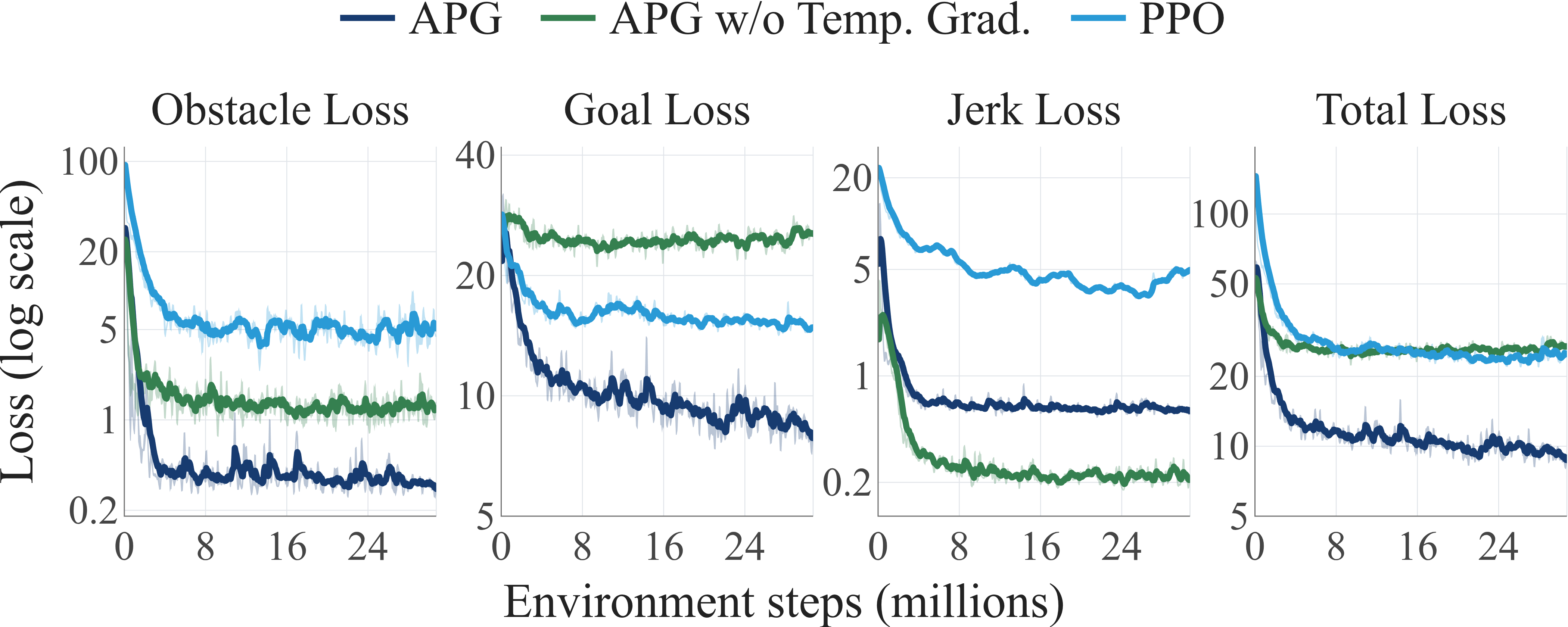}
    \caption{Task-loss curves during training for APG, APG without temporal gradient, and PPO. PPO uses the negative of the same task loss as APG as its training reward.}
    \label{fig:loss_curve}
    \vspace{-2mm}
\end{figure}

Figure~\ref{fig:loss_curve} shows that APG without temporal gradient exhibits low jerk loss and high goal loss despite receiving goal-loss gradients at every step. This suggests that neglecting the effects of current decisions on subsequent replanning steps favors local smoothness over sustained goal progress. PPO accounts for long-term outcomes through returns, but its policy gradients are noisy due to estimation from sampled rollouts. This gradient noise can lead to slower optimization, causing PPO to reduce the task loss more slowly and converge to a higher final loss than APG under the same training budget.

\begin{table}[t]
    \centering
    \caption{Comparison of policy optimization methods}
    \label{tab:ablation}
    \setlength{\tabcolsep}{5pt}
    \begin{tabular}{llcc}
        \toprule
        Map Density & Method
        & \shortstack{Success Rate}
        & \shortstack{Mean Speed (m/s)} \\
        \midrule
        \multirow{3}{*}{Sparse}
        & APG & \textbf{94\%} & \textbf{8.16} \\
        & \shortstack[l]{APG w/o Temp. Grad.}
        & \textbf{94\%} & 2.46 \\
        & PPO &60\% & 5.14 \\
        \midrule
        \multirow{3}{*}{Dense}
        & APG & \textbf{92\%} & \textbf{7.47} \\
        & \shortstack[l]{APG w/o Temp. Grad.}
        & 88\% & 2.05 \\
        & PPO & 64\% & 5.11 \\
        \bottomrule
    \end{tabular}
    \vspace{-2mm}
\end{table}

We evaluate each method without additional speed adjustment in the same sparse and dense environments used in Section~\ref{sec:benchmark}, reporting mean flight speed and success rate over 50 trials per density level. The learning characteristics observed during training are also reflected in flight performance. As shown in Table~\ref{tab:ablation}, APG without temporal gradient maintains high success rates but achieves mean flight speeds of only about $2$~m/s. PPO achieves higher mean flight speeds than APG without temporal gradient but has lower success rates. In contrast, full APG maintains high success rates while achieving mean flight speeds of around $7$--$8$~m/s.

\section{Real-World Experiments}
\label{sec:real_world}

We evaluate the proposed framework on physical robots using onboard sensing and state estimation. We first assess high-speed flight on a quadrotor platform and then apply the framework to an unmanned ground vehicle (UGV) to examine its extensibility to a different robotic platform.

\subsection{Quadrotor Flight Experiments}

\begin{figure}[t]
    \centering
    \includegraphics[
        width=0.98\linewidth,
        trim={540 40 530 40},
        clip
    ]{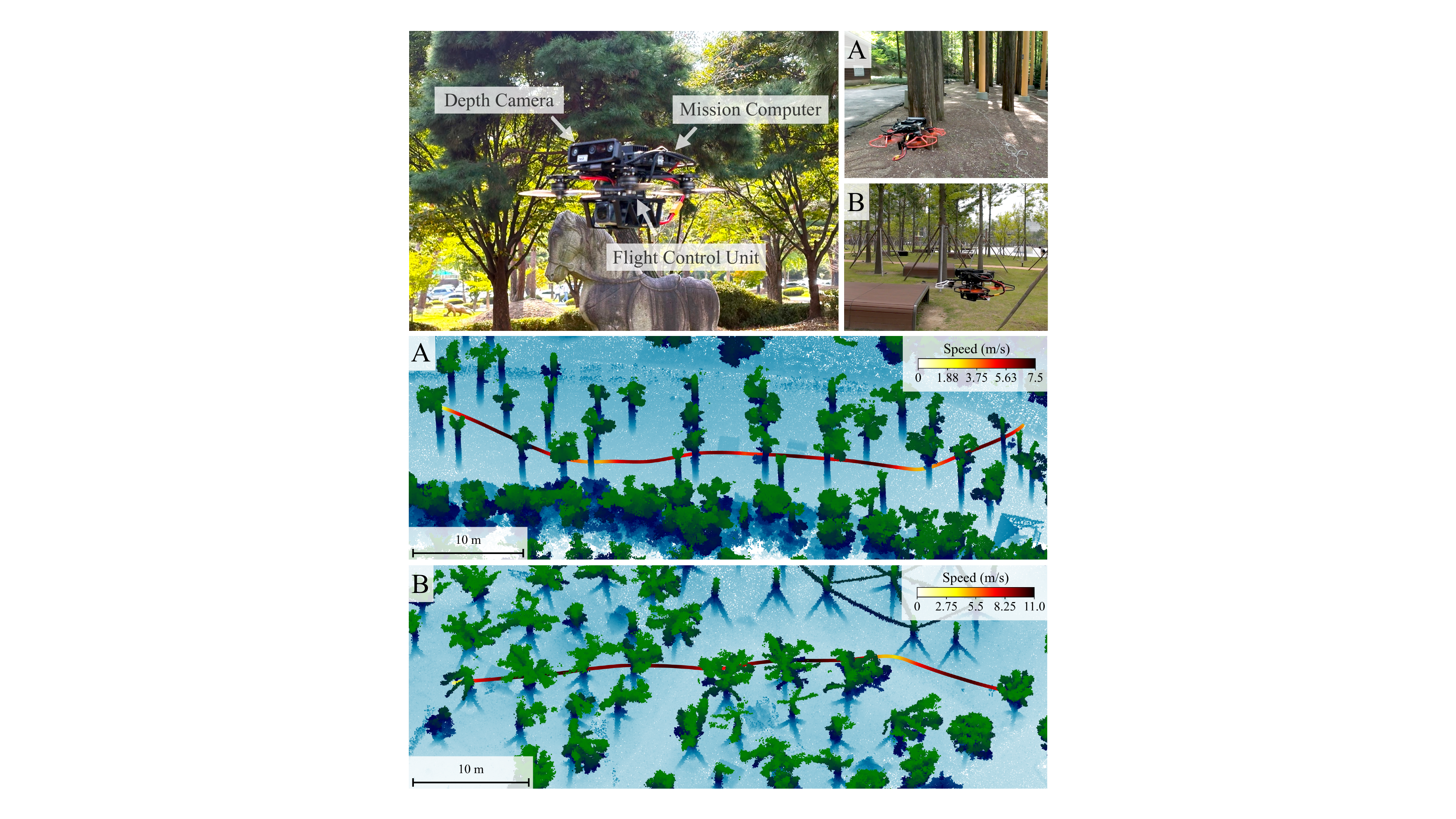}
    \caption{Experimental quadrotor platform and hardware configuration, together with flight results in diverse outdoor environments. The top-right images show two test environments (A and B), while the bottom plots visualize the corresponding flight trajectories, colored by speed.}
    \label{fig:quad_exp}
    \vspace{-2mm}
\end{figure}

We use the quadrotor platform shown in Fig.~\ref{fig:quad_exp}, with a diameter of approximately $25$~cm. The platform is equipped with an NVIDIA Jetson Orin NX for onboard computation, a Pixhawk 6C Mini flight controller for low-level flight control, and an OAK-D Pro Wide camera for depth perception. Its thrust-to-weight ratio is approximately $3.0$. The stereo camera provides an effective horizontal field of view of $97^{\circ}$ and a maximum depth sensing range of $15$~m.

We use OpenVINS-based visual-inertial odometry~\cite{openvins} for state estimation and a geometric controller for trajectory tracking. Depth images undergo the same preprocessing used during training. The acceleration input to the neural network is estimated from IMU measurements using an exponential moving average filter. The proposed planner performs planning at $10$~Hz, with neural network inference and trajectory generation taking approximately $13$~ms.

We conduct experiments at different flight speed settings in the two forest environments shown in Fig.~\ref{fig:quad_exp}. The proposed planner generates collision-free trajectories in both environments, allowing the quadrotor to safely avoid obstacles and reach the goal. The peak flight speed achieved in the experiments is approximately $11$~m/s.

Although training uses differential-flatness-based state transitions that assume accurate trajectory tracking, stable flight was achieved without extensive controller tuning for the proposed planner. The smoothness of the generated trajectories and receding-horizon replanning based on updated state estimates may have helped mitigate the effects of tracking errors.

\subsection{Extension to a Ground Robot}

Following~\cite{9592663}, we evaluate the proposed framework's extensibility to other differentially flat systems using the Clearpath Jackal, a skid-steer UGV, as a representative example. We model the ground robot as a unicycle and formulate it as a differentially flat system. The planar position serves as the flat output, and the heading is determined from the tangent direction of the planar trajectory. We also adjust the neural network output bounds to match the Jackal's hardware specifications. The learning framework and loss function remain the same as those used for the quadrotor.

For real-world experiments, we use a Jackal equipped with an Intel NUC for onboard computation. FAST-LIO2~\cite{xu2022fastlio2} provides state estimation using a Livox MID-360 LiDAR, while an OAK-D Pro Wide camera supplies depth observations to the planner. Trajectory planning runs onboard at $10$~Hz, with a simple PD controller for trajectory tracking. The system is evaluated in a real-world forest environment.

\begin{figure}[t]
    \centering
    \includegraphics[
        width=0.98\linewidth,
        trim={330 330 330 330},
        clip
    ]{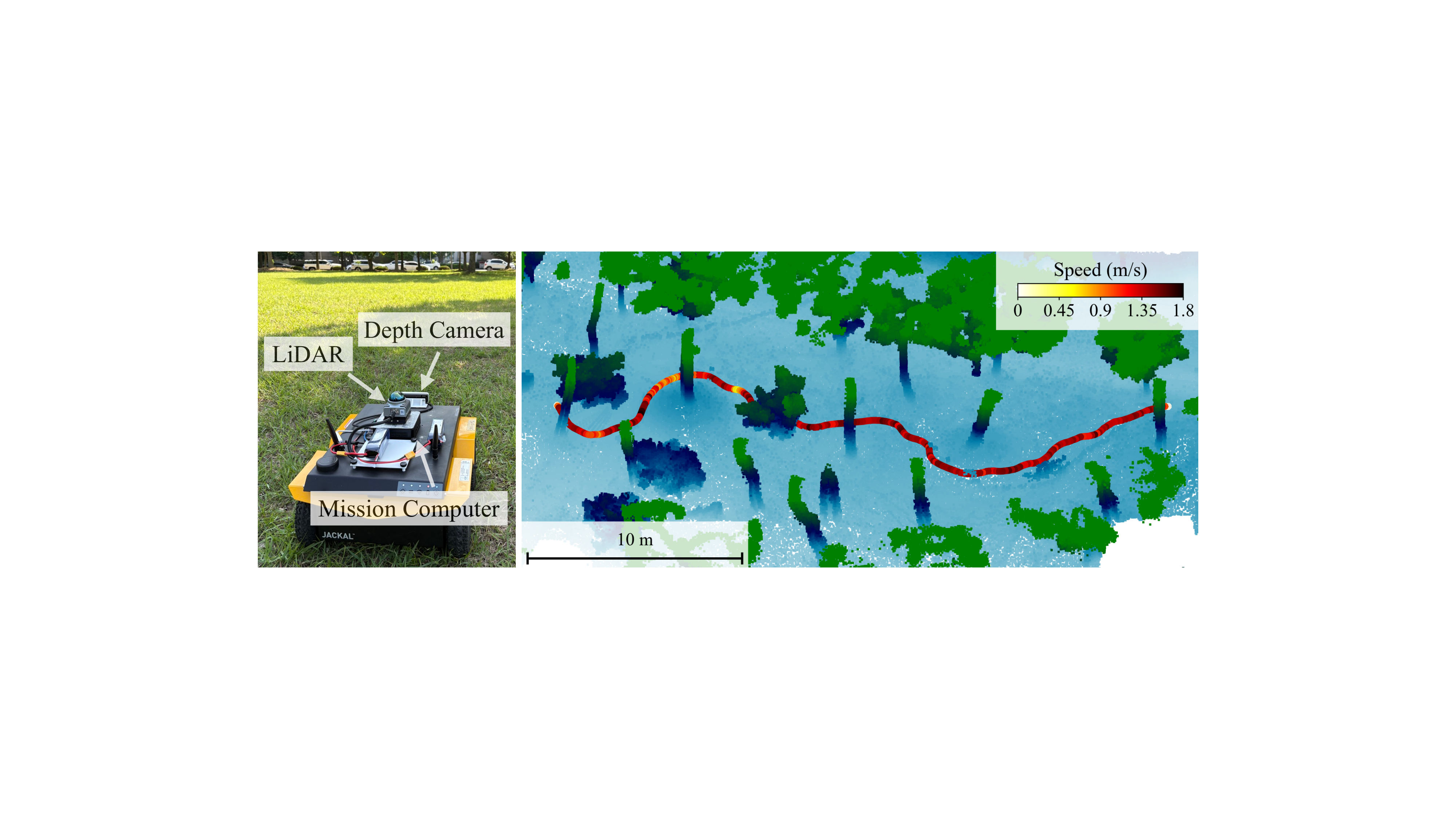}
    \caption{Hardware configuration of the Clearpath Jackal and visualization of real-world navigation experiments.}
    \label{fig:jackal_exp_new}
    \vspace{-4mm}
\end{figure}

As shown in Fig.~\ref{fig:jackal_exp_new}, the Jackal navigates around natural obstacles in the forest by generating and tracking collision-free trajectories. These results demonstrate that the proposed framework can be applied to other differentially flat robotic systems by adapting the flatness model and network output bounds to the target platform.

\section{Conclusion}
In this work, we presented APG-Planner, a vision-based high-speed trajectory planner trained with analytic policy gradients. By leveraging differential flatness, we construct differentiable trajectory-level state transitions that enable episode-level optimization across successive replanning decisions without explicitly unrolling low-level vehicle dynamics.

Extensive simulation and real-world experiments demonstrate that APG-Planner achieves comparable or superior planning performance to representative baselines and enables high-speed autonomous flight in diverse environments. We further transfer the framework to a skid-steer ground robot, demonstrating its applicability to other differentially flat robotic systems. These results highlight the effectiveness of combining analytic policy gradients with differential-flatness-based state transitions for learning online trajectory planners.

\bibliographystyle{IEEEtran}
\bibliography{refs}

\vfill
 
\end{document}